\documentclass[10pt,journal,compsoc]{IEEEtran}%
\usepackage{enumitem}
\usepackage{cite}
\usepackage{amsmath,amssymb,amsfonts}
\usepackage{algorithmic}
\usepackage{graphicx}
\usepackage{textcomp}
\usepackage{hyperref}
\usepackage{algorithm}
\usepackage[table,xcdraw]{xcolor}
\usepackage{comment}
\usepackage{amsthm}
\newtheorem{assumption}{Assumption}
\usepackage{titlesec}
\usepackage{makecell}
\usepackage{booktabs} 
\usepackage{multirow} 
\usepackage{caption}  
\usepackage{xcolor}

\usepackage[font=footnotesize,skip=0pt]{caption}
\titlespacing{\subsection}{0pt}{*0.6}{*0.5}
\def\BibTeX{{\rm B\kern-.05em{\sc i\kern-.025em b}\kern-.08em
    T\kern-.1667em\lower.7ex\hbox{E}\kern-.125emX}}
\begin{document}
\renewcommand{\paragraph}[1]{\textbf{#1}\mbox{}\\}


\title{H-FLTN: A Privacy-Preserving Hierarchical Framework for Electric Vehicle Spatio-Temporal Charge Prediction} 


\author{%
\IEEEauthorblockN{\IEEEauthorrefmark{1}Robert Marlin, \IEEEauthorrefmark{1}Raja Jurdak, Alsharif Abuadbba\IEEEauthorrefmark{2}} \\
%

\IEEEcompsocitemizethanks{\IEEEcompsocthanksitem \IEEEauthorrefmark{1}Robert Marlin, \IEEEauthorrefmark{1}Raja Jurdak with School of Computer Science, Queensland University of Technology, Australia. \IEEEauthorrefmark{1}Robert Marlin is also with CSIRO's Data61 and Cyber Security Cooperative Research Centre, Australia\{robert.marlin; r.jurdak\}@qut.edu.au.}
\IEEEcompsocitemizethanks{\IEEEcompsocthanksitem \IEEEauthorrefmark{2}Alsharif Abuadbba is with CSIRO's Data61, Australia\{sharif.abuadbba\}@data61.csiro.au.}
}

\markboth{H-FLTN: A Privacy-Preserving Hierarchical Framework for Electric Vehicle Spatio-Temporal Charge Prediction, February~2025}%
{Shell \MakeLowercase{\textit{et al.}}: Bare Demo of IEEEtran.cls for Computer Society Journals}

\maketitle

\begin{abstract}
The widespread adoption of Electric Vehicles (EVs) poses critical challenges for energy providers, particularly in predicting charging time (temporal prediction), ensuring user privacy, and managing resources efficiently in mobility-driven networks. This paper introduces the Hierarchical Federated Learning Transformer Network (H-FLTN) framework to address these challenges. H-FLTN employs a three-tier hierarchical architecture comprising EVs, community Distributed Energy Resource Management Systems (DERMS), and the Energy Provider Data Centre (EPDC) to enable accurate spatio-temporal predictions of EV charging needs while preserving privacy. Temporal prediction is enhanced using Transformer-based learning, capturing complex dependencies in charging behavior. Privacy is ensured through Secure Aggregation, Additive Secret Sharing, and Peer-to-Peer (P2P) Sharing with Augmentation, which allow only secret shares of model weights to be exchanged while securing all transmissions. To improve training efficiency and resource management, H-FLTN integrates Dynamic Client Capping Mechanism (DCCM) and Client Rotation Management (CRM), ensuring that training remains both computationally and temporally efficient as the number of participating EVs increases. DCCM optimises client participation by limiting excessive computational loads, while CRM balances training contributions across epochs, preventing imbalanced participation. Our simulation results based on large-scale empirical vehicle mobility data reveal that DCCM and CRM reduce the training time complexity with increasing EVs from linear to constant. By mitigating key FL challenges including data heterogeneity, computational overhead, and bias H-FLTN provides a secure, resource-efficient solution for predicting EV charging behavior. Its integration into real-world smart city infrastructure enhances energy demand forecasting, resource allocation, and grid stability, ensuring reliability and sustainability in future mobility ecosystems.
\end{abstract}



\section{\textbf{Introduction}} \label{sec:Introduction}

\par Global adoption of electric vehicles (EVs) has brought about significant changes in transportation systems and energy infrastructure, including demands and generation, with the promise of reduced greenhouse gas emissions and enhanced sustainability. However, this shift also poses significant challenges for energy providers, particularly in predicting and managing the spatial and temporal demand for EV charging. Addressing these challenges requires efficient resource management, and privacy-preserving solutions that can handle the dynamic and heterogeneous nature of EV mobility data, particularly the non-IID distributions of data generated by diverse EV usage patterns in smart city environments. One promising approach to addressing these complexities is Machine Learning (ML) using Federated Learning (FL), which enables distributed learning across decentralised systems while preserving data privacy.

\par Our earlier work in FL for EV charging concentrated on location prediction \cite{marlin2025privacypreservingchargelocation}, leaving the question of whether our model could be modified to predict an EV’s next charging time alongside location. This study explores a Transformer-based approach for dual-task predictions of both the location and time of the next EV charging event. The Transformer model is particularly well-suited for time prediction due to its ability to capture long-range dependencies through self-attention mechanisms. Unlike Convolutional Neural Networks (CNNs), which focus on spatial feature extraction, and Long Short-Term Memory (LSTM) networks, which process data sequentially with recurrent dependencies, Transformers capture entire sequences in parallel, enabling more efficient modeling of temporal patterns. This parallelism enables the model to learn complex temporal relationships more effectively, reducing the risk of vanishing gradients and preserving information over extended time horizons. By leveraging self-attention, the Transformer model dynamically weighs the importance of past time steps, making it highly effective for predicting future EV charging times in dynamic mobility environments. This solution evaluates continuous temporal predictions, addressing limitations in prior frameworks that primarily focused on categorical spatial outputs, such as discrete location classifications.

\par While FL enables privacy-preserving learning in decentralised EV networks, it faces scalability and bias challenges~\cite{kairouz2021advances}. Another study highlights that as client numbers grow, FL systems experience increased computational overhead, training latency, and resource consumption~\cite{li2020federated}. The proposed Hierarchical Federated Learning Transformer Network (H-FLTN) framework addresses these issues by integrating Dynamic Client Capping Mechanism (DCCM) and Client Rotation Management (CRM). DCCM optimises client participation per round, reducing computational burden without compromising generalisation, while CRM ensures balanced representation. By dynamically managing client selection and participation, these mechanisms enhance FL scalability by improving resource allocation and mitigating participation biases. This is particularly important in mobility-driven environments, where data is highly non-IID and uneven client contributions can distort model updates, ultimately reducing generalisation performance~\cite{vucinich2023current}. As the number of EVs participating in Hierarchical Federated Learning (HFL) increases, computational demands escalate, leading to longer training times and higher resource consumption. This challenge necessitates an efficient mechanism to balance computational efficiency with model accuracy. Tackling these challenges is crucial for robust and efficient large-scale FL training.

\par This study seeks to answer the following research question: \textit{"Can we develop a solution that ensures resource efficiency and user privacy while predicting an EV’s next charging time across diverse mobility patterns and battery capacities?"}

\par Accurately predicting an EV’s charging time is crucial for enhancing energy management in large-scale smart city ecosystems. Unlike location-based predictions, which focus on spatial demand, time prediction introduces additional complexities due to varying mobility behaviors and battery capacities. Addressing these temporal uncertainties requires efficient resource management and a privacy-preserving approach. In response to these challenges, this paper presents H-FLTN, a novel hierarchical learning framework that enables energy providers to anticipate demand patterns, optimise resource allocation, and improve overall grid stability through spatio-temporal predictions.

H-FLTN operates across three hierarchical levels: individual EVs as clients, community Distributed Energy Resource Management Systems (DERMS), and the Energy Provider Data Centre (EPDC). By sharing only secret shares, which represent model weights, between EVs and the community DERMS, the framework preserves user privacy while enabling accurate predictions at the DERMS level. These predictions, encompassing location and time metrics, are securely transmitted to the EPDC for detailed analysis using secure communications. The EPDC leverages this data to assess energy demands across regions, assisting energy providers in optimising energy distribution and resource management.

\par The proposed framework addresses key challenges in FL through the following contributions:

\begin{enumerate} 
\item \textbf{Optimisation of Client Participation via Dynamic Client Capping Mechanism (DCCM):} Optimises client participation in HFL by dynamically capping active clients per training round to achieve maximum efficiency. This reduces computational overhead while maintaining model accuracy, leading to faster convergence, reduced training time, and improved learning efficiency.

\item \textbf{Balanced Participation with Client Rotation Management (CRM):} Promotes balanced client participation in HFL by systematically rotating clients within capped subsets across training epochs. This approach mitigates participation bias, prevents skewed model updates, and enhances generalisation, resulting in a more robust and adaptable model.

\item \textbf{Integration of a Privacy-Preserving Hierarchical Federated Learning Transformer Network (H-FLTN):} Proposal of a three-tier hierarchical framework for predicting spatial and temporal EV charging behaviours, incorporating privacy-preserving methods and resource management strategies to enable robust decentralised learning while ensuring data protection.

\item \textbf{Extensive Evaluation and Performance Analysis:} Comprehensive evaluation of the H-FLTN framework through rigorous simulations to assess predictive accuracy, resource management efficiency, and bias mitigation, validating its suitability for large-scale EV mobility applications.
\end{enumerate}

\par These contributions collectively enable the proposed H-FLTN framework to address the critical challenges of privacy and scaleable resource management in decentralised learning. By ensuring secure and unbiased participation, the framework facilitates accurate spatio-temporal predictions, fostering trust among users and advancing the integration of EVs into sustainable urban systems.

\par The remainder of this paper is organised as follows. Section~\ref{sec:Related Work} details Related Works in Modeling Approaches for Spatio-temporal Predictions; including Bidirectional Long Short-Term Memory (BiLSTM), CNN, Transformers, and HFL; addressing privacy, scaleable resource management, and accuracy. Section~\ref{sec:ProblemDefinition} formalises our H-FLTN EV Next Charge Prediction Problem. Section~\ref{sec:MathematicalFramework} introduces the model's architecture and optimisation process. Section~\ref{sec:Proposed Solution} discusses our approach to Privacy-Preserving EV Next Charge Location and Time Prediction, including our novel application of privacy features within the framework. Section~\ref{sec:Performance Evaluation} evaluates our proposed hierarchical approach to this problem area, Baseline Modelling, and Results and Analysis. Section~\ref{sec:Discussion} discusses any possible trade-offs. Lastly, we conclude our paper with a discussion on future work.


\section{\textbf{Related Work}} \label{sec:Related Work}

\par FL enables distributed model training while maintaining data privacy by avoiding direct data sharing. H-FL extends this concept by introducing multi-level aggregation, improving scalability and adaptability in dynamic environments like EV networks \cite{yang2019federated}, and \cite{liu2020client}. Despite these advancements, challenges persist in scaleable resource management, ensuring privacy, and mitigating adversarial attacks in large-scale, real-world systems.

\par Several studies have proposed techniques to address these challenges. Lalitha et al. \cite{lalitha2019peer} introduced P2P sharing as a decentralised collaboration mechanism, and Shamir et al. \cite{shamir1979share} proposed Additive Secret Sharing to enhance privacy during model updates. Bonawitz et al. \cite{bonawitz2016practical} demonstrated the use of Secure Aggregation to protect individual contributions during weight aggregation, while McMahan et al. \cite{mcmahan2017communication} explored strategies to balance client participation and ensure fairness in decentralised systems. While effective, these methods often address specific challenges in isolation. The proposed H-FLTN framework builds on and integrates approaches to create a unified architecture for privacy-preserving, accurate spatio-temporal prediction, and optimally managed resource allocation in mobility-driven environments.

\subsection{\textbf{Modeling Approaches for Spatio-Temporal Predictions}}

\par Various machine learning models have been explored for spatio-temporal prediction tasks. Yan et al. \cite{yan2025attention} highlighted BiLSTM's strengths in capturing sequential dependencies but noted its limitations in computational efficiency and scalability. Marlin et al. \cite{marlin2024electric} showed that CNNs effectively capture spatial dependencies but struggle with variable input dimensions and computational overhead.

\par Transformer models, with their multi-head attention mechanisms, have emerged as a superior choice for spatio-temporal prediction \cite{vaswani2017attention}. By attending to both spatial and temporal dependencies, Transformers capture complex relationships in EV mobility data, such as location and charging time interactions. Marlin et al. \cite{marlin2025privacypreservingchargelocation} demonstrated that Transformers outperformed BiLSTM and CNN architectures in predicting EV charging locations across datasets of varying sizes.

\par This study extends prior work by introducing a dual-task framework for predicting both the time and location of the next EV charging event. Unlike approaches focused solely on categorical outputs (EV next charge location)~\cite{marlin2025privacypreservingchargelocation}, this method leverages Mean Squared Error (MSE) for continuous temporal predictions, improving precision. By addressing key challenges in modeling dynamic mobility patterns, this study provides energy providers with actionable insights into both spatial and temporal energy demands.

\subsection{\textbf{Hierarchical Federated Learning (H-FL)}}

\par H-FL extends traditional FL by introducing hierarchical layers for multi-level aggregation, enabling scalability and adaptability in heterogeneous data environments. Li et al. \cite{li2020federated} demonstrated H-FL's potential for managing data heterogeneity, while Yang et al. \cite{yang2019federated} and Liu et al. \cite{liu2020client} showcased its effectiveness in dynamic real-world scenarios. To address challenges in privacy researchers have proposed techniques such as P2P Sharing and Augmentation and Secure Aggregation. The H-FLTN framework builds on these foundations, integrating privacy-preserving mechanisms and efficient resource allocation strategies to enhance utility for spatio-temporal prediction tasks in EV networks.

\par Despite H-FL's scalability, ensuring efficient client participation while managing resource and time constraints remains an open challenge. Existing works \cite{kairouz2021advances, smith2017federated} explore FL optimisation but largely overlook computational load balancing in hierarchical settings. While H-FL distributes computation and communication load, optimising client selection for balanced compute expenditure and training efficiency remains critical for scalability.

\par To address this, H-FLTN introduces Dynamic Client Capping Mechanism (DCCM) and Client Rotation Management (CRM), extending prior work on client selection \cite{cho2020client}. DCCM dynamically regulates participation, reducing compute usage and training delays~\ref{tab:epoch_times}, while CRM ensures systematic client rotation, mitigating selection bias and enabling diverse client participation. By splitting training into smaller, more efficient allocations, these mechanisms accelerate convergence, improving accuracy in fewer iterations while lowering compute expense.

\par A key contribution of this research is a novel time prediction approach within H-FLTN. Prior studies on FL for EV charging \cite{marlin2025privacypreservingchargelocation} primarily focus on location-based forecasting, while time prediction remains largely unexplored. Unlike location, which benefits from stable infrastructure, time prediction is influenced by dynamic factors such as congestion and grid constraints. By training on temporal features from historical client data, our model learns recurring patterns, allowing EV owners to schedule charging more efficiently while enabling energy providers to anticipate demand across locations and times for proactive resource allocation.

\subsection{\textbf{Integration of Privacy and Efficient Resource Management Techniques}}

\par The H-FLTN framework integrates multiple privacy-preserving techniques within a hierarchical structure to ensure data confidentiality. P2P Sharing, proposed by Lalitha et al. \cite{lalitha2019peer}, enables decentralised collaboration among EVs while maintaining privacy. Additive Secret Sharing \cite{shamir1979share} ensures model updates remain confidential, while Secure Aggregation \cite{bonawitz2016practical} prevents adversaries from reconstructing individual EV contributions. TLS 1.3 and MeLSeC further secure communication channels, mitigating risks of interception or unauthorised access.

\par The combination of DCCM (client capping) and CRM (client rotation) enhances resource management and addresses participation bias. Unlike the single-level sampling strategies of McMahan et al. \cite{mcmahan2017communication}, DCCM improves training efficiency in hierarchical systems by selectively engaging a subset of clients per round, reducing training overhead, shortening epoch durations~\ref{tab:epoch_times}, and accelerating convergence.

\par Wang et al. \cite{wang2019adaptive} explored adaptive client selection to optimise resource utilisation in FL, highlighting the need to balance computational constraints with model accuracy. Expanding on this, DCCM dynamically adjusts participation for efficient training without compromising performance. When combined with CRM, it maintains balanced participation across training rounds, preventing selection bias while preserving learning quality.

\section{\textbf{Problem Definition}} \label{sec:ProblemDefinition}

\par Predicting the next charging location and time for EVs is a critical challenge for managing energy demand in dynamic EV networks. This paper focuses on time prediction, which is novel in our work, while also addressing location prediction, which remains crucial for efficient energy management. Additionally, safeguarding user privacy is an integral aspect of this problem, as sensitive EV data is vulnerable to various types of attacks in decentralised networks as discussed in studies by Chen et al.~\cite{chen2022data} and Khowaja et al.~\cite{khowaja2023spin}. Accurate spatio-temporal predictions, combined with robust privacy mechanisms, enable energy providers to optimise charging infrastructure, manage peak demand periods, and balance energy loads across regions.

\par Traditional centralised approaches to predictive modelling struggle with the complexity of EV networks due to the following limitations:

\begin{itemize}
    \item \textbf{Scalability Challenges}: Large-scale EV network solutions must address issues such as high communication costs, varying client availability, and computational load imbalances. DCCM and CRM mitigate these by dynamically managing client participation, reducing unnecessary computation, and ensuring balanced training across rounds.
    
    \item \textbf{Privacy Risks}: Transferring sensitive EV data to central servers increases the risk of data breaches, raising significant privacy concerns.

    \item \textbf{Dynamic and Heterogeneous Data}: EV mobility patterns exhibit high variability due to differences in user behaviour, and regional charging infrastructure. This heterogeneity extends beyond spatiotemporal mobility data to include variations in charging frequency, energy consumption patterns, and EV types, all of which complicate predictive modelling.

    \item \textbf{Threat Model and Security Considerations}: FL in decentralised EV networks introduces privacy and security risks due to the distributed nature of data and model updates. Our threat model assumes an honest-but-curious adversary attempting to extract information from shared updates. The primary security concern in this study is ensuring that model updates remain confidential and are not linkable to specific EV clients.

    \item \textbf{Privacy-Preserving Mechanisms}: To protect client confidentiality, the H-FLTN framework employs the following mitigation strategies:
    
    \begin{enumerate}[label=\alph*)]
        \item \textbf{Non-transitory EVs}: \textit{P2P Sharing and Augmentation for Secret Shares} disperses each update among 2–10 peers when available. If fewer than two peers are available, the EV sends its encrypted secret shares directly to the DERMS instead. This decentralised approach ensures that no single EV’s raw model update can be reconstructed, maintaining data confidentiality.
    
        \item \textbf{Transitory EVs}: These EVs submit their secret shares directly to the community DERMS. Secure Aggregation ensures that individual contributions remain private and cannot be traced back to a specific EV, reinforcing confidentiality.
    \end{enumerate}

    \item \textbf{Outlier Mitigation}:  The community DERMS applies normalisation to aggregated weights, reducing the influence of extreme values. While this helps smooth anomalies, it does not fully eliminate adversarial influence. However, in combination with secure aggregation, normalisation reduces the impact of outlier updates before they contribute to the final model update.

    \item While privacy threats such as inference and data linkage attacks are the primary focus of this study, securing model integrity against other known attacks is left for future work.  
\end{itemize}

\par By focusing on these core threats, our H-FLTN enhances privacy and security while maintaining the efficiency of model training in EV networks.

\par These vulnerabilities highlight the need for robust privacy-preserving methods, particularly in the context of distributed prediction tasks. Addressing these risks is essential to ensuring user trust while maintaining accurate and efficiently managed predictions of both location and time. Specifically, the model must account for:
\begin{itemize}
    \item \textbf{Current State Features}: The EV’s current battery level \(E_t\) and location \(l_t\).
    \item \textbf{Historical Patterns}: Previous mobility data \(H_{1:t}\), such as past pick-up and drop-off locations, distances travelled, and timestamps.
    \item \textbf{Spatial Relationships}: Neighbouring locations \(B(l_t)\) that are geographically or operationally connected to the current location.
    \item \textbf{Temporal Factors}: Charging times and usage patterns, expressed as \(T_t\), which are critical for understanding and predicting temporal trends.
\end{itemize}

\vspace{3mm}
\hrule\vspace{1mm}

\par The following notations define key variables used in the H-FLTN framework:  
\vspace{1mm}\hrule\vspace{1mm}
\begin{itemize}
    \item \(L\): The set of all possible locations, where \(l_i \in L\) represents an EV’s location.
    \item \(B(l_t)\): The set of neighbouring locations that share a border with \(l_t\).
    \item \(E_t\): The remaining EV battery charge at time \(t\), expressed as a percentage (0\% to 100\%).
    \item \(H_{1:t}\): The historical mobility data, including distances travelled, previous locations, and timestamps.
    \item \(T_t\): The recorded timestamp of an EV charge event.
    \item \( \hat{L}_t \): The predicted next charging location for the EV.
    \item \( \hat{T}_t \): The predicted time of the next charge.
\end{itemize}

\par The objective is to develop a function \(f\) that predicts an EV’s next charging location \(\hat{n}_t\) and time \(\hat{T}_n\) based on the following inputs: 
\begin{equation}
    \hat{n}_t, \hat{T}_t = f(E_t, l_t, B(l_t), T_t, H_{1:t}).
\end{equation}
\par Here, \(E_t\), \(l_t\), \(B(l_t)\), and \(T_t\) represent the current state of the EV, while \(H_{1:t}\) provides historical mobility data to improve predictive accuracy.

\par The problem is subject to the following constraints:
\begin{itemize}
    \item Predictions are only made if the EV’s battery level is sufficient for another trip (\(E_t > 20\%\)).
    \item If \(E_t \leq 20\%\), the EV’s current location \(l_t\) is assumed to be the next charging location, and charging is expected immediately (\(\hat{T}_n = T_t\)).
    \item Charging station availability affects predictions, meaning the next charging location \(\hat{n}_t\) must belong to the set of available stations \(S_t\), where \(\hat{n}_t \in S_t\).
\end{itemize}

\par This leads to the following conditional equation:
\begin{equation}
    (\hat{n}_t, \hat{T}_t) =
    \begin{cases} 
        f(E_t, H_{1:t}, l_t, B(l_t), T_t) & \text{if } E_t > 20\% \\
        (l_t, T_t) & \text{otherwise}.
    \end{cases}
\end{equation}

\par The non-linear, multidimensional nature of this prediction task necessitates advanced ML techniques. FL is employed to collaboratively train models across decentralised EV data, ensuring raw data remains local and private. By leveraging FL and incorporating features such as P2P Sharing and Augmentation, Additive Secret Sharing, Secure Aggregation, and encrypted communication protocols including TLS 1.3 and MeLSeC, this study enhances data confidentiality and ensures data integrity while preserving privacy These methods enhance model robustness while maintaining the trade-offs between predictive accuracy, resource management, and privacy, enabling accurate spatio-temporal predictions in large-scale EV networks.

\vspace{2mm}
\hrule


\section{Mathematical Framework} \label{sec:MathematicalFramework}

\noindent This section presents the formalisation of our \textit{H-FLTN}, detailing the mathematical constructs governing local model training, P2P Sharing and Augmentation with Additive Secret Sharing, Secure Aggregation, DCCM \& CRM, and Prediction Transmission to EPDC. The formulation ensures privacy-preserving, resource-scalable, and efficient model updates in decentralised EV networks. Unix timestamp features are used as inputs to predict the next EV charging time, enabling the model to process time as a continuous variable rather than classifying it into categories. This enables the model to capture fine-grained temporal patterns, allowing the model to achieve accurate charge-time predictions.

\subsection{Local Model Training and P2P Sharing with Additive Secret Sharing}
Let \(\mathcal{D}_i\) represent the local dataset for EV \(i\), where \(i \in \{1, 2, \dots, N\}\) and \(N\) is the total number of EVs. Each EV \(i\) trains a local model \(f_i(W_i^t)\) on its dataset \(\mathcal{D}_i\), where \(W_i^t\) represents the model parameters (weights) for EV \(i\) at time \(t\). 

\par The goal of local training is to find the optimal model parameters \(W_i^t\) by minimising the local loss function \(\mathcal{L}_i\):
\begin{equation}
    W_i^t = \arg \min_{W} \mathcal{L}_i(f_i(W), \mathcal{D}_i)
\end{equation}
where:
\begin{itemize}
    \item \( W_i^t \) represents the optimised local model weights for EV \( i \) at time \( t \).
    \item \( W \) is the set of all possible model parameters being optimised.
    \item \( \mathcal{L}_i \) is the local loss function for EV \( i \).
\end{itemize}

\par After local training:

\begin{assumption}
\label{assumption:peer_count}
Given the rapid increase in EV adoption, we assume that a minimum of two non-transitory EVs per community will be available for P2P sharing. This assumption is supported by projections indicating that by 2050, over 70\% of global vehicle sales will be EVs, ensuring sufficient density for this approach~\cite{EnerOutlook21}.
\end{assumption}

\begin{itemize} 
    \item Non-transitory EVs engage in P2P sharing to enhance privacy and robustness. Once the locally trained weights \(W_i^t\) are partitioned into secret shares \(S_{i,k}^t\), these secret shares are shared within a limited peer group of 2–10 EVs. Each EV receives secret shares from its peers and aggregates them to compute augmented secret shares \(\tilde{S}_i^t\), where augmentation introduces controlled modifications to prevent direct inference of local weights while preserving the integrity of the global model.  
    \item Peer selection is dynamic and proximity-based, considering:
        \begin{itemize}
            \item Proximity-based selection (favoring EVs operating in similar regions to reflect shared energy demand characteristics).
            \item Availability in the current training round.
            \item Rotation prevents biased weight sharing. If at least 2 EVs are available, the system proceeds with P2P Sharing. If fewer than 2 EVs are available, the EV submits encrypted shares directly to the community DERMS, ensuring uninterrupted training.
        \end{itemize}
    \item EVs communicate via secure channels using MeLSeC to negotiate and exchange secret shares.
\end{itemize}

\begin{equation}
    \tilde{W}_i^t = W_i^t + \alpha \sum_{j \in \text{peers}} S_j^t
\end{equation}
Where \(\alpha\) is a scaling factor controlling the contribution of peer updates.

\par Transitory EVs, by contrast, directly partition their locally trained weights into secret shares:
\begin{equation}
    S_{i,k}^t = \text{Partition}(W_i^t), \quad W_i^t = \sum_{k=1}^{K} S_{i,k}^t
\end{equation}

\par These secret shares are transmitted to the community DERMS, ensuring that local model weights remain undisclosed.  
\begin{assumption}
\label{assumption:honest_but_curious}
While community DERMS receives all secret shares from transitory EVs, community DERMS is considered honest-but-curious, following protocols without attempting weight reconstruction. Secure Aggregation ensures that only the final aggregated model is revealed, preventing DERMS from isolating individual EV contributions.
\end{assumption}

\subsection{Additive Secret Sharing and Secure Aggregation}
Each EV \(i\) partitions its weights into secret shares:
\begin{equation}
    S_{i,k}^t = \text{Partition}(W_i^t), \quad W_i^t = \sum_{k=1}^{K} S_{i,k}^t
\end{equation}
For non-transitory EVs, secret shares are distributed among 2–10 peers and augmented with their own shares before being sent to the community DERMS for secure aggregation. In contrast, transitory EVs transmit their secret shares directly to the community DERMS without peer distribution.

At the community DERMS, shares are aggregated without reconstruction of local weights:
\begin{equation}
    \Theta = \frac{1}{N} \sum_{i=1}^{N} S_i^t, \quad S_i^t = \sum_{k=1}^{K} S_{i,k}^t
\end{equation}

To mitigate privacy risks at DERMS:
\begin{itemize}
    \item \textbf{MeLSeC encryption}: Secures secret shares during transmission.
    \item \textbf{Normalisation}: Reduces outlier influence:
    \begin{equation}
        \tilde{\Theta} = \text{Normalise}(\Theta)
    \end{equation}
    \item \textbf{Peer-based Distribution}: Further obfuscates updates before DERMS aggregation.
\end{itemize}

\subsection{Dynamic Client Capping and Rotation}
To optimise computational efficiency, DCCM limits the number of active EV clients to \(C\), where \(C \leq N\). The CRM mechanism ensures clients are rotated across training rounds:

\begin{equation}
    \mathcal{C}_t = \text{Rotate}(\{1, 2, \dots, N\}, t, C)
\end{equation}

By capping active clients at \(C\), aggregation complexity is reduced from:
\begin{equation}
    \mathcal{O}(N) \quad \text{to} \quad \mathcal{O}(C), \quad C \ll N
\end{equation}

\subsection{Prediction Transmission to EPDC}
After aggregation, DERMS generate spatio-temporal predictions:
\begin{equation}
    \hat{n}_t, \hat{T}_n = \text{DERMS}(\tilde{\Theta})
\end{equation}
Where:
\begin{itemize}
    \item \( \hat{n}_t \) is the predicted next charging location.
    \item \( \hat{T}_n \) is the predicted charging time.
\end{itemize}

Predictions are securely transmitted to EPDC using TLS 1.3 for energy distribution analysis.

\subsection{Federated Learning Process}
The FL process integrates P2P Sharing and Augmentation, Additive Secret Sharing, Secure Aggregation, DCCM, CRM, and Normalisation:

\begin{algorithm}[H]
\caption{FL Process for H-FLTN with DERMS and EPDC in EV Charging Networks}
\begin{algorithmic}[1]
\STATE Initialise global model weights \(\Theta^0\)
\FOR {each round \(t = 1, 2, \dots\)}
    \STATE Select capped and rotated client set: \(\mathcal{C}_t = \text{Rotate}(\{1, \dots, N\}, t, C)\)
    \FOR {each EV \(i \in \mathcal{C}_t\) \textbf{in parallel}}
        \STATE Receive global model weights: \( W_i^t = \Theta^t \)
        \STATE Train local model \( f_i(W_i^t) \) on local dataset \( \mathcal{D}_i \)
        \IF{EV \(i\) is non-transitory}
            \STATE Partition local weights into secret shares: \( S_{i,k}^t = \text{Partition}(W_i^t) \)
            \STATE Determine available peers: \( P_i^t = \text{SelectPeers}(i, \text{min}(2, \text{AvailablePeers}(i))) \)
            \IF{\(|P_i^t| > 1\)}
                \STATE Exchange secret shares with peers in \( P_i^t \) for augmentation
                \STATE Transmit augmented secret shares to DERMS
            \ELSE
                \STATE Transmit encrypted secret shares directly to DERMS
            \ENDIF
        \ELSE 
            \STATE Partition local weights into secret shares: \( S_{i}^t = \text{Partition}(W_i^t) \)
            \STATE Transmit secret shares directly to DERMS
        \ENDIF
    \ENDFOR
    \STATE Aggregate all received secret shares at DERMS: \( S_{\text{agg}}^t = \sum_{i=1}^{N} S_i^t \)
    \STATE Reconstruct global weights from aggregated secret shares: \( \tilde{\Theta} = \text{Reconstruct}(S_{\text{agg}}^t) \)
    \STATE Normalise global weights: \( \tilde{\Theta} = \text{Normalise}(\tilde{\Theta}) \)
    \STATE Update global model: \( \Theta^{t+1} = \tilde{\Theta} \)
    \STATE Send updated global weights back to local clients: \( W_i^{t+1} = \Theta^{t+1}, \forall i \in \mathcal{C}_t \)
    \STATE DERMS generates predictions: \( (\hat{n}_t, \hat{T}_n) = \text{DERMS}(\Theta^{t+1}) \)
    \STATE Transmit predictions to EPDC using TLS 1.3
\ENDFOR
\end{algorithmic}
\end{algorithm}

\par The H-FLTN framework ensures accurate and private spatio-temporal predictions in large-scale, mobility-driven EV networks.


\section{\textbf{Proposed Solution}} \label{sec:Proposed Solution}

\par This section presents the: 1) the solution architecture, 2) the EV mobility dataset, and 3) the proposed H-FLTN system. First, we outline the H-FLTN pipeline, detailing the sequence of processes from data input to the EPDC's logistics output. Second, we describe the construction and characteristics of the EV mobility dataset, focusing on the methodologies used for data collection, feature engineering, and the assumptions embedded in the dataset. Finally, we discuss the implementation of our H-FLTN system, elaborating on its structure, training process, and specific application in predicting the EV's next charging location and time while preserving user privacy.


\subsection{\textbf{Proposed Pipeline}} \label{subsec:Pipeline_Diagram}

\par
\begin{figure}[h!]
    \centering
    \caption{H-FLTN Pipeline figure}
    \vspace{10pt}
    \includegraphics[scale=0.41]{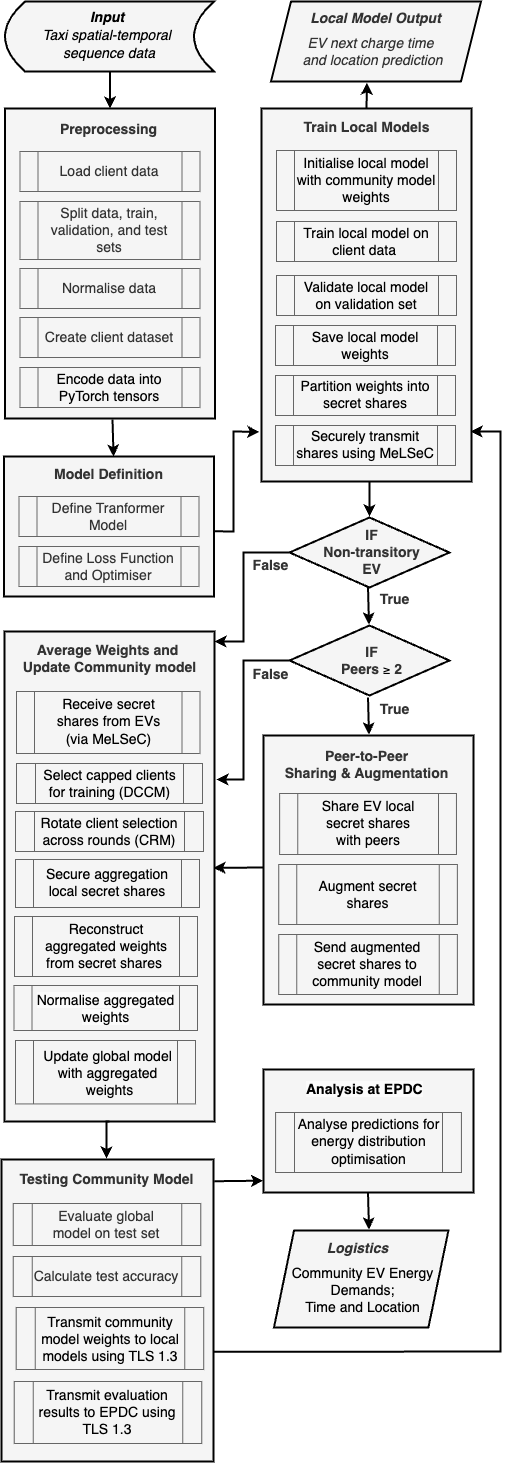}
    \par\vspace{10pt}
    \footnote*{Pipeline diagram illustrating end-to-end processes of the H-FLTN framework}
    \vspace{5pt}
    \label{fig:Pipeline_Diagram}
\end{figure}

 The pipeline illustrated in Figure~\ref{fig:Pipeline_Diagram} consists of the following \textbf{key stages}:  

\begin{enumerate}  
    \item \textit{Preprocessing}:  
    Each EV processes its historical spatio-temporal mobility data independently, ensuring privacy. The data is split into training, validation, and test sets, normalised, and converted into PyTorch tensors for efficient model training.  

    \item \textit{Local Model Training and Prediction}:  
    \begin{itemize}  
        \item EV clients initialise their models with community global model weights.  
        \item Models are trained locally using EV-specific charging behaviour data.  
        \item Once trained, the local model outputs its prediction for the EV’s next charge time and location.  
        \item Local model weights are then partitioned into secret shares for secure transmission.  
    \end{itemize}  

    \item \textit{Peer-to-Peer (P2P) Sharing and Augmentation}:  
    \begin{itemize}  
        \item Non-transitory EVs securely exchange and augment secret shares with 2–10 community peers before transmitting them to the DERMS via MeLSeC. If fewer than 2 peers are available, the EV transmits its encrypted secret shares directly to the DERMS instead.         
        \item Transitory EVs bypass P2P sharing and send their secret shares directly to the community DERMS.  
    \end{itemize}  

    \item \textit{Community Model Aggregation at DERMS}:  
    \begin{itemize}  
        \item The DERMS module receives encrypted secret shares from participating EVs.  
        \item Client participation is regulated through the Dynamic Client Capping Mechanism (DCCM) and Client Rotation Management (CRM), ensuring a balanced and efficient training process.  
        \item Secure aggregation using additive secret sharing is applied to compute the global model weights without exposing individual client contributions.  
        \item Normalisation is performed to mitigate outliers before updating the community global model.  
    \end{itemize}  

    \item \textit{Testing and Global Model Update}:  
    \begin{itemize}  
        \item The updated global model is evaluated on a test set to assess prediction accuracy for location and charge time.  
        \item If validated, the finalised weights are securely distributed back to local EV models via TLS 1.3 for the next training round.  
    \end{itemize}  

    \item \textit{Prediction Output and EPDC Analysis}:  
    \begin{itemize}  
        \item Each local model retains its prediction for the EV’s next charge time and location.  
        \item Additionally, model outputs are transmitted to the Energy Provider Data Centre (EPDC), where they support energy demand forecasting and grid optimisation.  
    \end{itemize}  
\end{enumerate}  

\subsection{\textbf{EV Mobility Dataset}} \label{subsec:Dataset}

\par To develop and validate H-FLTN, we required a dataset containing detailed EV mobility and charging patterns. For this study, we utilised a modified version of the Chicago taxi mobility dataset \cite{chicago2016}, enriched with synthetic data to incorporate EV specific metrics. While originally based on non-EV taxi mobility, these enhancements reflect future adoption trends, assuming that:

\begin{assumption}
\label{assumption:ev_transition}
The majority of taxi fleets will transition to EVs by 2050.
\end{assumption}

\begin{assumption}
\label{assumption:charging_infrastructure}
Uniform charging infrastructure will be in place, providing consistent charging speeds and compatibility.
\end{assumption}

\par To simulate realistic EV behaviour, the dataset includes nine EV models with battery capacities ranging from 143 km to 416 km, randomly assigned to reflect fleet diversity. Trips span one year across 77 distinct community areas, representing potential pick-up, drop-off, or charge locations, forming the study’s spatial framework. Battery charge levels range from 20\% to 100\%, aligning with industry standards for preserving battery health \cite{batteryhealth}, ensuring diverse charging scenarios from immediate to long-term needs.

\par Each EV serves as a client in the H-FLTN system, contributing to decentralised learning. The framework is designed to optimise resource handling, ensuring scalability and adaptability across different urban settings, provided sufficient spatio-temporal data is available. This flexibility highlights H-FLTN's applicability in efficiently managing EV energy demands across diverse smart city environments.

\subsection {\textbf{Training}} \label{subsec:Training}

\par The proposed H-FLTN framework employed ML on spatio-temporal datasets representing EV trips, with each EV treated as an independent client within the system's hierarchical structure. The datasets preserved the chronological order of sequences to maintain the temporal dependencies essential for accurate location and time predictions. Each EV client trained model weights locally, which were then partitioned into secret shares and securely aggregated at the community-level DERMS. The DERMS predictions were subsequently transmitted to the EPDC for city-wide spatio-temporal EV energy demand analysis.

\par Experiments utilised dataset groups organised by EV numbers: 150, 300, 500, 750, and 1000. Each group was further divided into three subgroups (A, B, and C), representing diverse community-level compositions of EVs. These subgroups were randomly selected but designed to reflect real-world variability in EV distributions across different regions, where factors such as battery capacity, energy demand, and mobility patterns differ across communities. This structure allows for evaluating the robustness of the model across heterogeneous EV populations, ensuring that results are generalisable beyond a single distribution of EVs. Data within each group was split into training, validation, and test sets, with 85\% allocated for training and validation and the remaining 15\% for testing. Within the training and validation subset, an additional 85-15\% split was applied to maintain representative validation sets. This stratified approach ensured balanced representation of mobility patterns across training phases while preserving chronological order to maintain spatio-temporal dependencies.

\par The random assignment of EVs to subgroups inherently created non-IID (non-independent and identically distributed) data distributions, as each subgroup captured a diversified combination of EV models, battery capacities, mobility patterns, and charging behaviours. While traditional methods often simulate non-IID data using a Dirichlet distribution controlled by an \(\alpha\) parameter where smaller \(\alpha\) values increase data heterogeneity our approach achieves a similar outcome by leveraging random distributions. This method ensures the model learns from a broad range of real-world scenarios and generalises effectively to dynamic environments.

\par To ensure the framework aligned with real-world use cases, datasets represented a wide range of battery charge levels, from 20\% to 100\%. This diversity allowed the framework to capture varying charging behaviours, enabling short-term predictions for low battery levels (e.g., 20\%) and long-term predictions for higher charge levels (e.g., 100\%). This alignment with real-world charging patterns enhanced the model's utility for energy providers, supporting both short-term and long-term resource planning.

\renewcommand{\arraystretch}{1.2} 
\setlength{\tabcolsep}{8pt}      
\begin{table}[h!]
\centering 
\caption{Central Model Results Table.}
\vspace{4mm}
\label{tab: CentralModelResultsTable}
\begin{tabular}{|l|l|l|l|}
\hline
\textbf{}                               & \textbf{BiLSTM}         & \textbf{CNN}            & \textbf{Transformer}    \\ \hline
\rowcolor[HTML]{EFEFEF} 
{\color[HTML]{333333} \textbf{100 EVS}} & {\color[HTML]{333333} } & {\color[HTML]{333333} } & {\color[HTML]{333333} } \\ \hline
\textbf{Group A}                   & 56.25\%                 & 50.00\%                 & 97.07\%                 \\ \hline
\textbf{Group B}                   & 47.78\%                 & 48.88\%                 & 98.43\%                 \\ \hline
\textbf{Group C}                   & 48.82\%                 & 49.62\%                 & 95.65\%                 \\ \hline
\textbf{}                               &                         &                         &                         \\ \hline
\rowcolor[HTML]{EFEFEF} 
\textbf{200 EVS}                        &                         &                         &                         \\ \hline
\textbf{Group A}                   & 43.00\%                 & 46.23\%                 & 97.49\%                 \\ \hline
\textbf{Group B}                   & 46.00\%                 & 44.29\%                 & 98.35\%                 \\ \hline
\textbf{Group C}                   & 45.89\%                 & 42.89\%                 & 97.29\%                 \\ \hline
\textbf{}                               &                         &                         &                         \\ \hline
\rowcolor[HTML]{EFEFEF} 
\textbf{300 EVS}                        &                         &                         &                         \\ \hline
\textbf{Group A}                   & 46.68\%                 & 48.25\%                 & 98.95\%                 \\ \hline
\textbf{Group B}                   & 44.79\%                 & 43.78\%                 & 97.58\%                 \\ \hline
\textbf{Group C}                   & 43.92\%                 & 48.27\%                 & 97.89\%                 \\ \hline
\end{tabular}\vspace{10pt}\\
\footnotesize{Table 1 displays centralised model outcomes for the three ML architectures: CNN, BiLSTM, and Transformer. These ML models were evaluated using datasets containing EV numbers including 100, 200, and 300, which were further divided into three subgroups containing mixed EV models covering a range of remaining battery capacities from 20\% to 100\%.}
\end{table}


\section {\textbf {Performance Evaluation}} \label{sec:Performance Evaluation}

\par This section evaluates the effectiveness of the proposed H-FLTN framework in predicting both the location and time of EVs’ next charging event. We compare centralised and decentralised ML modelling approaches and analyse the results of the decentralised H-FLTN framework. Particular focus is given to the framework's ability to maintain high time and location prediction accuracy.

\subsection {\textbf {Baseline Modelling}} \label{subsec:Baseline Modelling}

\par The following section includes baseline experiments, where we evaluated three centralised ML models: BiLSTM, CNN, and a Transformer. These models were tested on spatio-temporal datasets to determine their ability to interpret EV sequences that include both mobility and charge transaction features. The results, presented in Table:~\ref{tab: CentralModelResultsTable}, confirmed that the Transformer model demonstrated superior performance in capturing spatio-temporal dependencies, making it the optimal choice for further decentralised experiments.


\renewcommand{\arraystretch}{1.2} 
\setlength{\tabcolsep}{8pt}      
\begin{table}[h!]
\centering 
\caption{Comparison of H-FL (BiLSTM) and H-FLTN Accuracy (\%).}
\vspace{4mm}
\label{tab:HFL_vs_HFLTN}
\begin{tabular}{|l|l|l|}
\hline
\textbf{}                               & \textbf{H-FL (BiLSTM)}  & \textbf{H-FLTN}         \\ \hline
\rowcolor[HTML]{EFEFEF} 
\textbf{100 EVS}                         &                         &                         \\ \hline
\textbf{Group A}                         & 89.16\%                 & 97.25\%                 \\ \hline
\textbf{Group B}                         & 89.03\%                 & 96.05\%                 \\ \hline
\textbf{Group C}                         & 88.89\%                 & 97.95\%                 \\ \hline
\textbf{}                               &                         &                         \\ \hline
\rowcolor[HTML]{EFEFEF} 
\textbf{200 EVS}                         &                         &                         \\ \hline
\textbf{Group A}                         & 87.01\%                 & 97.06\%                 \\ \hline
\textbf{Group B}                         & 87.25\%                 & 96.99\%                 \\ \hline
\textbf{Group C}                         & 89.25\%                 & 97.08\%                 \\ \hline
\textbf{}                               &                         &                         \\ \hline
\rowcolor[HTML]{EFEFEF} 
\textbf{300 EVS}                        &                         &                         \\ \hline
\textbf{Group A}                         & 87.08\%                 & 97.90\%                 \\ \hline
\textbf{Group B}                         & 86.30\%                 & 98.10\%                 \\ \hline
\textbf{Group C}                         & 87.49\%                 & 97.07\%                 \\ \hline
\end{tabular}\vspace{10pt}\\
\footnotesize{Table compares the performance of H-FL (BiLSTM) and H-FLTN for EV charging prediction. The evaluation was conducted on datasets containing 100, 200, and 300 EVs, divided into three subgroups (A, B, and C).}
\end{table}

\par Building on these findings, we compare H-FL (BiLSTM) with H-FLTN. The results presented in Table~\ref{tab:HFL_vs_HFLTN} demonstrate a significant performance improvement of the H-FLTN model over the H-FL (BiLSTM) achieving an average accuracy increase of approximately 8-11\%. This improvement is most pronounced in the 100 EVS dataset, where H-FLTN reaches a peak accuracy of 97.95\% in Group C, compared to 88.89\% for H-FL (BiLSTM). Similarly, for larger datasets such as 200 EVS and 300 EVS, H-FLTN maintains high accuracy, surpassing 97\% in all cases, whereas H-FL (BiLSTM) exhibits a gradual decline in performance, with the lowest accuracy recorded at 86.30\% for 300 EVS - Group B.

\par The superior performance of H-FLTN can be attributed to the advantages of the Transformer architecture, which replaces the recurrent nature of BiLSTM with a self-attention mechanism. Unlike BiLSTM, which processes sequences sequentially and retains past dependencies through hidden states, the Transformer model leverages multi-head self-attention to capture long-range dependencies more effectively. This allows H-FLTN to model complex spatio-temporal relationships in EV charging patterns without suffering from vanishing gradients or the constraints of sequential processing. 

\par Testing for all modelling was conducted on a 2022 MacBook Pro equipped with an M1 Max CPU, 64 GB of memory, and a 2 TB SSD. The system runs macOS Sequoia, providing a robust environment for handling the computational demands of our modeling and simulation experiments. All experiments were executed using Visual Studio Code (Version 1.97.0) as the primary development environment, which facilitated efficient code management and debugging.

\subsection {\textbf {Results and Analysis}} \label{subsec:Results and Analysis}

\par This section discusses and analyses predictive results for EV next charge prediction including both location and time metrics. The performance of the H-FLTN framework was systematically evaluated across multiple EV group sizes, with the results for the location prediction presented in Table~\ref{tab:location_model_results} and the results for the time prediction in Table~\ref{tab:time_model_mse_results}. 

\par The evaluation of the H-FLTN framework involved datasets representing different sizes of EV groups, specifically 150, 300, 500, 750, and 1000 EVs. To ensure a comprehensive and unbiased evaluation, each group was further divided into three subgroups A, B, and C. This stratification prevents evaluation bias by mitigating potential data imbalances that could arise from a single large, undivided dataset. The EVs in each group and subgroup were selected in a completely random fashion, ensuring a natural distribution of variability in battery capacities, mobility patterns, and charging behaviors. This setup mirrors real-world conditions, where EV behaviors are inherently heterogeneous. By evaluating across multiple diverse subgroups, the framework’s robustness, bias mitigation, and ability to generalise across varying mobility and charging conditions can be effectively assessed.

\subsubsection {\textbf {Location and Time Prediction Results}} \label{subsubsec:Location Prediction Results}

\par This section presents the location prediction results, analysing the effectiveness of the H-FLTN framework in capturing spatial dependencies across diverse EV groups. Table~\ref{tab:location_model_results} provides a performance breakdown across different group sizes, highlighting trends in accuracy. This analysis examines how factors like battery charge capacity distributions and mobility patterns affect the model's ability to generalise. By examining these factors, we evaluate how well the framework maintains accurate predictions across diverse datasets and dynamic charging behaviors.

\captionsetup{skip=10pt}
\begin{table}[h!]
\centering 
\caption{Location Model Results}
\begin{tabular}{|l|c|c|c|c|}
\hline
\textbf{EVs} & \textbf{Subgroup} & \textbf{Max (\%)} & \textbf{Med (\%)} & \textbf{SD} \\
\hline
\multirow{4}{*}{150} & A & 97.25 & 95.13 & 0.52 \\
                     & B & 96.05 & 95.96 & 0.44 \\
                     & C & 97.95 & 95.91 & 0.26 \\
\hline
\textbf{Overall}     & - & 97.95 & 95.91 & 0.42 \\
\hline\hline
\multirow{4}{*}{300} & A & 95.66 & 94.57 & 2.28 \\
                     & B & 97.15 & 96.43 & 1.21 \\
                     & C & 97.50 & 96.30 & 0.60 \\
\hline
\textbf{Overall}     & - & 97.50 & 96.30 & 1.53 \\
\hline\hline
\multirow{4}{*}{500} & A & 97.06 & 94.72 & 5.42 \\
                     & B & 96.99 & 95.50 & 1.21 \\
                     & C & 97.08 & 96.21 & 1.37 \\
\hline
\textbf{Overall}     & - & 97.08 & 95.50 & 3.30 \\
\hline\hline
\multirow{4}{*}{750} & A & 97.03 & 96.44 & 2.49 \\
                     & B & 96.48 & 95.77 & 0.57 \\
                     & C & 97.91 & 96.29 & 0.47 \\
\hline
\textbf{Overall}     & - & 97.91 & 96.29 & 1.49 \\
\hline\hline
\multirow{4}{*}{1000} & A & 97.90 & 96.89 & 1.87 \\
                      & B & 98.10 & 97.30 & 0.86 \\
                      & C & 97.07 & 96.36 & 0.56 \\
\hline
\textbf{Overall}      & - & 98.10 & 96.89 & 1.23 \\
\hline
\end{tabular}
\label{tab:location_model_results}
\par\vspace{10pt} 
\footnotesize{This location model results table summarises the prediction performance across different groups of EVs. Metrics include maximum accuracy (Max, \%), median accuracy (Med, \%), and standard deviation (SD), providing insights into the variability, and highest observed values across subgroups. Overall represents a central measure encompassing all three subgroups.}
\end{table}

\begin{table}[h!]
\centering
\caption{Time Model MSE Results}
\begin{tabular}{|l|c|c|c|}
\hline
\textbf{EVs} & \textbf{Subgroup} & \textbf{Best MSE} & \textbf{Median MSE} \\
\hline
\multirow{4}{*}{150} & A & 0.5598 & 0.9862 \\
                     & B & 0.0933 & 0.0168 \\
                     & C & 0.0821 & 0.3314 \\
\hline
\textbf{Overall}     & - & 0.0821 & 0.6007 \\
\hline\hline
\multirow{4}{*}{300} & A & 0.0627 & 0.0287 \\
                     & B & 0.2215 & 0.8018 \\
                     & C & 0.0572 & 0.0115 \\
\hline
\textbf{Overall}     & - & 0.0572 & 0.4632 \\
\hline\hline
\multirow{4}{*}{500} & A & 0.0311 & 0.6612 \\
                     & B & 0.1849 & 0.8776 \\
                     & C & 0.0552 & 0.1852 \\
\hline
\textbf{Overall}     & - & 0.0311 & 0.6433 \\
\hline\hline
\multirow{4}{*}{750} & A & 0.4213 & 0.7437 \\
                     & B & 0.1149 & 0.5258 \\
                     & C & 0.0734 & 0.2265 \\
\hline
\textbf{Overall}     & - & 0.0734 & 0.5418 \\
\hline\hline
\multirow{4}{*}{1000} & A & 0.2471 & 0.8143 \\
                      & B & 0.1840 & 0.4221 \\
                      & C & 0.0841 & 0.7414 \\
\hline
\textbf{Overall}      & - & 0.0841 & 0.6809 \\
\hline
\end{tabular}
\par\vspace{5pt} 
\footnotesize{This table presents the MSE results for the time prediction model across different EV groups (150, 300, 500, 750, and 1000 EVs). The Best MSE column represents the lowest recorded error for each subgroup, while the Median MSE provides a more robust central measure, reducing the impact of outliers. Each EV group is divided into subgroups A, B, and C, with an overall performance metric reported for each group. These results provide insights into model accuracy and stability across varying dataset sizes.}
\label{tab:time_model_mse_results}
\end{table}

\par The 'Overall' metrics in Table~\ref{tab:location_model_results} summarise the results across all subgroups. The overall maximum accuracy (\textbf{Max (\%)}) reflects the highest accuracy observed among all subgroups. The overall median accuracy (\textbf{Med (\%)}) is the median value of the subgroup medians, providing a central tendency metric. The overall standard deviation (\textbf{SD}) is calculated using the pooled standard deviation formula, accounting for variability across all subgroups while considering their respective sample sizes.  

\par The location prediction results in Table~\ref{tab:location_model_results} demonstrate the effectiveness of the H-FLTN framework in capturing spatial dependencies across diverse EV groups. The model achieved a median accuracy of 95.91\% for the smallest EV group (150 EVs) and 96.89\% for the largest group (1000 EVs), with similarly strong performance across intermediate sizes. Notably, the highest accuracy observed was 98.10\%, achieved in Group B of the 1000 EV set. The relatively low standard deviation of 1.23 for the largest EV group further reinforces the framework's stability and reliability, even in scenarios with significant data heterogeneity.

\par While some variation is present, the pooled SD values indicate that the model maintains consistency across different EV groups. Analysis of Group B in the 1000 EV set revealed that this subgroup contained a higher distribution of EVs with lower battery capacities and low remaining battery charge. This led to a greater number of recorded mobility patterns, as these EVs required more frequent charging, resulting in a denser and more diverse dataset of spatio-temporal charging behaviors. This alignment reduced variability within the data and enhanced the model’s ability to generalise effectively during training.

\par The time prediction results, detailed in Table~\ref{tab:time_model_mse_results}, demonstrate the model’s ability to provide accurate and actionable temporal insights. To evaluate accuracy, we use the Mean Squared Error (MSE), a widely adopted metric for regression models that quantifies the average squared difference between predicted and actual values. A lower MSE indicates higher accuracy, as it signifies a smaller deviation from the true time values. 

\par To improve interpretability, we compare MSE values to approximate time intervals to assess the model’s ability to predict the next EV charge event. On average, the model demonstrates strong predictive performance across all EV groups, with corresponding MSE values validating its ability to make precise time predictions. For example, in the 150EV group, the model achieves a best-case MSE of 0.0821, which corresponds to an error margin within a few hours of the actual charging event. As the dataset scales, the best-case MSE remains consistently low, with the lowest best-case MSE recorded at 0.0311, aligning with an error within approximately an hour. This reinforces the model’s effectiveness across different dataset sizes. While median error slightly increases in larger datasets, the model provides useful temporal forecasting capabilities, suggesting that further refinements could improve resolution in finer time intervals.

\par The model utilises a Unix timestamp-based approach, where the predicted output is represented as an absolute time value. This timestamp corresponds to the conversion of predictions into interpretable month, day, and hour components. This method ensures a continuous temporal representation, allowing the model to successfully capture charging patterns.

\par Our test MSE results represent the overall error for the model’s time predictions evaluated on the test set. By leveraging Unix time as an absolute measure, the model effectively learns patterns in EV charging behaviour over various temporal scales. Results for Group A exhibit higher MSE values compared to Groups B and C, which can be attributed to greater variability in charging behaviours within this subgroup. The random assignment of EVs resulted in Group A containing a larger proportion of sequences with irregular charging patterns, making precise predictions more challenging. Despite this, the overall accuracy remains strong, these findings highlight the capability of our H-FLTN model to understand spatio-temporal mobility data.


\subsection {\textbf {Ablation Study}} \label{subsec:Ablation Study}

\par The Ablation Study provides an in-depth analysis of individual contributions and interplay of key components in the H-FLTN framework. By isolating specific mechanisms and evaluating their impact, this section underscores the significance of features such as Capping and Rotation, P2P Sharing with Augmentation, Additive Secret Sharing, and Secure Aggregation. This section is segmented into three key areas: Compute and Time Resource Optimisation, an analysis of client capping  and rotating (DCCM +RCM), Location Modelling, which assesses the framework’s ability to predict EV charging locations; and Time Modelling, which evaluates temporal predictions for EV charging. Together, these experiments demonstrate how each feature affects the H-FLTN framework across diverse use cases.

\subsubsection{\textbf{Compute and Time Resource Optimisation (DCCM + RCM)}} \label{subsubsec: Compute and Time Resource Optimisation}

\par To manage computational and time overheads, we implement client capping and rotating mechanisms DCCM + CRM, dynamically selecting an optimal number of clients based on extensive experimentation. We calculate the computational cost of modelling with H-FLTN using FLOPs (Floating Point Operations), considering both the forward and backward passes through the model architecture, as outlined by Tang et al.\cite{tang2018flops}. The total FLOPs per global epoch is computed as the sum of FLOPs across all participating clients, making it directly proportional to the number of clients involved in each epoch. 

\par Findings from these experiments highlight the effectiveness of the DCCM + CRM mechanisms in managing resource demands. When client participation is capped at 150 EVs, computational cost remains consistent across varying dataset sizes, reflecting that the cap, rather than the total number of available clients, dictates resource usage. FLOPs for both the 500 EV and 1000 EV groups yield identical values of 74,880,000 per epoch under DCCM + CRM. Without capping, the FLOPs scale directly with client count, doubling from 249,600,000 for 500 EVs to 499,200,000 for 1000 EVs. These findings reinforce the role of DCCM + CRM in optimising compute resources to maintain efficiency as network scale grows.

\begin{table}[h]
    \centering
    \caption{DCCM \& CRM Epoch Time Analysis for EV Groups}
    \begin{tabular}{|c|c|c|}
        \hline
        \textbf{EV Group} & \textbf{No (DCCM + CRM)} & \textbf{Using (DCCM + CRM)} \\
        \hline
        1000 EV & 1:51:23 & 0:16:28 \\
        750 EV  & 1:13:52 & 0:16:18 \\
        500 EV  & 0:58:00 & 0:19:46 \\
        300 EV  & 0:31:10 & 0:17:22 \\
        150 EV  & 0:18:03 & 0:18:45 \\
        \hline
    \end{tabular}
    \par\vspace{10pt} 
    \footnotesize{Epoch times (HH:MM:SS) for different EV group sizes, comparing training without and with DCCM + CRM. These mechanisms, applied at the DERMS level, optimise client participation and load balancing, reducing computational and time overheads.}
    \label{tab:epoch_times}
\end{table}

\begin{figure}[h!]
    \centering
    \caption{DCCM + CRM Line Plot}
    \includegraphics[width=9.2cm]{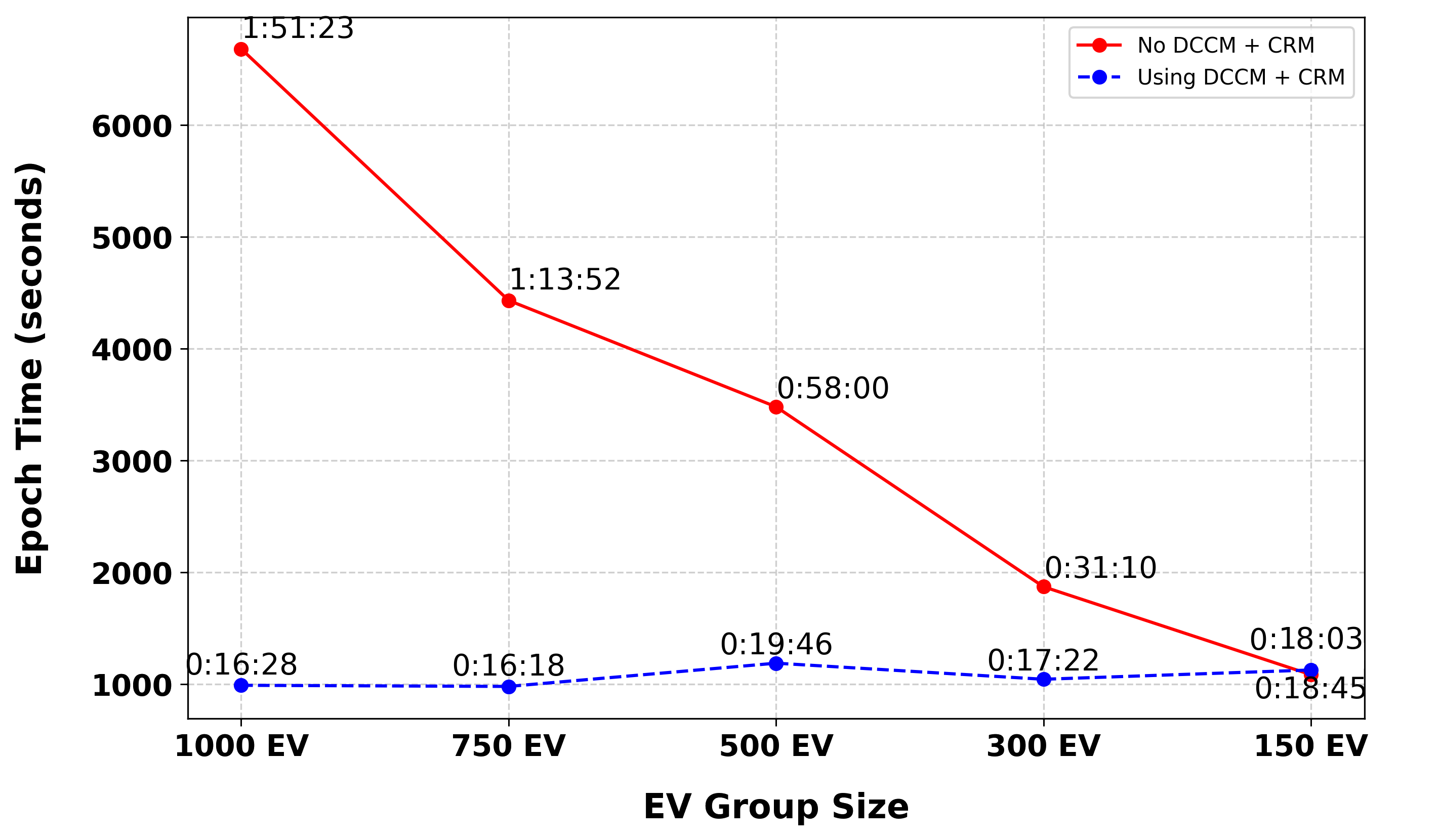}
    \par\vspace{10pt}
    \footnotesize{Epoch training time for different EV group sizes with and without DCCM and CRM. The X axis represents EV group size, and the Y axis represents epoch time in seconds, line labels are in HH:MM:SS format. The red line shows training without these mechanisms, while the blue dashed line represents training using DCCM and CRM. }
    \vspace{10pt}
    \label{fig:DCCM + CRM Line Plot}
\end{figure}

\par DCCM and CRM enhances model performance by dynamically adjusting the number of participating EVs per epoch to an optimal subset. This reduces training time and computational cost, allowing for faster convergence to the loss function minimum. In the 1000 EV group, training time decreased from 1:51:23 without DCCM + CRM to 0:16:28 with their application, while the 500 EV group saw a reduction from 0:58:00 to 0:19:46, demonstrating enhanced efficiency in optimising training time expenses.

\par To evaluate the impact of CRM on model performance and overfitting within our H-FLTN framework, we ran experiments under two conditions: (1) using CRM to rotate client selections per epoch and (2) not using CRM, where client selections stayed the same across epochs. We focused on three key metrics; Per-Epoch Diversity Ratio, Cumulative Diversity Ratio, and Training Loss — to assess client diversity, bias mitigation, and learning efficiency.

\par When CRM wasn’t used, the client selection process kept engaging the same subset of clients across all epochs. This resulted in a Per-Epoch Diversity Ratio of 0.3000 and a flat Cumulative Diversity Ratio of 0.3000. The lack of client diversity caused the model to overfit quickly to the limited data it kept seeing. The Training Loss Decrease Rate was minimal, dropping by only 0.04\%, showing that the model wasn’t making meaningful learning progress. More critically, the Generalisation Gap the difference between the training and test loss hit 3\% by the final epoch. This highlighted the model’s struggle to generalise beyond the overexposed data subset, confirming overfitting.

\par In contrast, when CRM was used, it introduced new clients in each epoch, ensuring that no client was reused. While the Per-Epoch Diversity Ratio remained at 0.3000, the Cumulative Diversity Ratio steadily increased over time, eventually reaching 0.9000 indicating that client selection maintained a consistent level of diversity. This broader exposure to varied data sources improved the model’s learning path. The Training Loss Decrease Rate was 12.7\%, indicating that the model was genuinely learning and adapting to new information throughout the training process. More importantly, this learning progress didn’t lead to overfitting. The Generalisation Gap shrank to 1.35\%, confirming that the model could generalise well to unseen data while maintaining efficient learning.

\subsubsection {\textbf {Location Prediction Ablation Study}} \label{subsubsec: Location Prediction Ablation Study}

\par The ablation study evaluates the contribution of key features Capping \& Rotating, Secret Sharing, Secure Aggregation, and Normalisation on the performance of our 500EV experimental group for our location model. The results are presented in Table~\ref{tab:location_model_ablation_study}, showing the test accuracy and standard deviation (SD) for subgroups A, B, and C, as well as overall metrics. For comparison, the base model results (Table~\ref{tab:location_model_results}) achieved an overall median accuracy of 95.50\% with an SD of 3.30. Subgroup median accuracies were 94.72\% (A), 95.50\% (B), and 96.21\% (C).

\begin{table}[h!]
\centering
\caption{Location Model Ablation Study Results for our 500EV Group}
\begin{tabular}{|c|c|c|c|}
\hline
\textbf{Feature Removed} & \textbf{Group} & \textbf{Test Acc (\%)} & \textbf{SD} \\
\hline
Capping \& Rotating & A & 94.26 & 1.37 \\
        500EV        & B & 93.50 & 1.21 \\
                     & C & 93.21 & 1.37 \\
\hline
\textbf{Overall}     & - & 94.26 & 1.32 \\
\hline\hline
Secret Sharing    & A & 94.18 & 1.15 \\
      500EV         & B & 92.91 & 1.22 \\
                     & C & 93.83 & 1.12 \\
\hline
\textbf{Overall}     & - & 94.18 & 1.16 \\
\hline\hline
Secure Aggregation & A & 92.36 & 1.37 \\
       500EV         & B & 92.41 & 1.21 \\
                     & C & 91.88 & 1.32 \\
\hline
\textbf{Overall}     & - & 92.41 & 1.30 \\
\hline\hline
Normalisation     & A & 83.84 & 1.15 \\
       500EV         & B & 83.17 & 1.02 \\
                     & C & 83.97 & 1.12 \\
\hline
\textbf{Overall}     & - & 83.97 & 1.10 \\
\hline
\end{tabular}
\label{tab:location_model_ablation_study}
\par\vspace{10pt} 
\footnotesize{This location model ablation table summarises the results for our 500EV experimental group, under various ablation settings. Metrics include test accuracy (Test Acc) and standard deviation (SD) for groups A, B, and C, highlighting the impact of removing specific features on the model's performance.}
\end{table}

\subsubsection{Feature Analysis}\label{subsubsec: Feature Analysis}

\par Removing individual features revealed their distinct contributions to model performance:

\begin{itemize}
    \item \textbf{Capping \& Rotating:} Removing this feature caused the overall test accuracy to drop to 94.26\% (SD: 1.32). Subgroup accuracies also decreased to 94.26\% (A), 93.50\% (B), and 93.21\% (C). This demonstrates that while Capping \& Rotating enhances efficiency and fairness, its removal has a moderate impact on accuracy.
    \item \textbf{Secret Sharing:} The exclusion of Secret Sharing reduced overall accuracy to 94.18\% (SD: 1.16), with subgroup accuracies of 94.18\% (A), 92.91\% (B), and 93.83\% (C). The feature's minimal impact on accuracy underscores its utility in enhancing privacy without compromising performance.
    \item \textbf{Secure Aggregation:} Removing Secure Aggregation caused a more significant decline in overall accuracy to 92.41\% (SD: 1.30), with subgroup accuracies of 92.36\% (A), 92.41\% (B), and 91.88\% (C). This feature is crucial for robust weight integration, as its absence reduces performance.
    \item \textbf{Normalisation:} The removal of Normalisation resulted in the most significant drop in overall accuracy to 83.97\% (SD: 1.10). Subgroup accuracies fell to 83.84\% (A), 83.17\% (B), and 83.97\% (C). Normalisation's absence severely disrupts gradient flow and convergence, directly impacting stability and accuracy.
\end{itemize}

\par The study highlights the balanced contributions of individual features to the FL framework. Features like Capping \& Rotating reduces computational expense, while Secret Sharing enhances privacy, whereas Secure Aggregation and Normalisation are vital for optimising performance and stability. Together, these features create an efficient, resource-optimised framework, ensuring robustness and adaptability for real-world large-scale deployments.

\subsubsection {\textbf {Time Prediction Ablation Study}} \label{subsubsec: Time Prediction Ablation Study}

\par The results in Table \ref{tab:time_model_ablation_study} highlight the critical role of specific features in achieving optimal performance and efficiency for the time model. 

\begin{table}[h!]
\centering
\caption{Time Model Ablation Study Results for our 500EV Group}
\begin{tabular}{|c|c|c|c|}
\hline
\textbf{Feature Removed} & \textbf{Group} & \textbf{Best MSE} & \textbf{Median MSE} \\
\hline
Capping \& Rotating & A & 0.0335 & 0.6741 \\
        500EV        & B & 0.1927 & 0.8894 \\
                     & C & 0.0583 & 0.1936 \\
\hline
\textbf{Overall}     & - & 0.0335 & 0.6539 \\
\hline\hline
Secret Sharing    & A & 0.0324 & 0.6678 \\
      500EV         & B & 0.1872 & 0.8821 \\
                     & C & 0.0561 & 0.1894 \\
\hline
\textbf{Overall}     & - & 0.0324 & 0.6481 \\
\hline\hline
Secure Aggregation & A & 0.0341 & 0.6793 \\
       500EV         & B & 0.1955 & 0.8923 \\
                     & C & 0.0602 & 0.1981 \\
\hline
\textbf{Overall}     & - & 0.0341 & 0.6575 \\
\hline\hline
Normalisation     & A & 0.0328 & 0.6710 \\
       500EV         & B & 0.1894 & 0.8852 \\
                     & C & 0.0570 & 0.1918 \\
\hline
\textbf{Overall}     & - & 0.0328 & 0.6508 \\
\hline
\end{tabular}
\label{tab:time_model_ablation_study}
\par\vspace{10pt} 
\footnotesize{This time model ablation table summarises the results for our 500EV experimental group under various ablation settings. Metrics include Best MSE and Median MSE for groups A, B, and C, highlighting the minor impact of removing specific features on model performance.}
\end{table}

\par The results of the time model ablation study, presented in Table~\ref{tab:time_model_ablation_study}, demonstrate the effect of removing key features including; Capping and Rotating, Secret Sharing, Secure Aggregation, and Normalisation on the predictive performance of the model. The observed changes in MSE values are minimal, confirming that these features primarily influence computational efficiency, privacy, and client bias mitigation rather than the core predictive accuracy for our H-FLTN  model.

\par The removal of Capping and Rotating resulted in a minor increase in MSE values, with the best MSE shifting from 0.0311 to 0.0335 and the median MSE slightly increasing across all subgroups. This outcome aligns with expectations, as capping and rotating mechanisms regulate client participation to optimise resource allocation and mitigate bias, rather than directly influencing the model’s temporal learning capability.

\par Secret sharing and Secure Aggregation, which are implemented for privacy preservation, exhibited similar trends. The best MSE increased marginally to 0.0324 and 0.0341, respectively, while median values showed minor fluctuations. Since these techniques operate during the secure exchange and aggregation process rather than modifying the learning process itself, their removal had little effect on the model's ability to capture temporal dependencies.

\par The removal of Normalisation also led to negligible changes in predictive performance. The best MSE increased slightly to 0.0328, while the median MSE remained within a close range of prior results. This aligns with theoretical expectations, as the model utilises Unix timestamps, which inherently encode temporal sequences in a way that does not require Normalisation for stability.

\par Overall, the ablation study confirms that while these features are essential for real-world deployment in FL systems  ensuring privacy, bias mitigation, and efficiency their removal does not significantly impact the model's predictive accuracy. This is expected, as the transformer-based architecture leverage self-attention mechanisms to capture long-range dependencies across the entire temporal sequence simultaneously. This design allows the model to effectively encode time-based relationships without requiring explicit feature scaling, sequential processing constraints, or handcrafted temporal smoothing. As a result, the predictive performance remains stable, even when auxiliary mechanisms such as Normalisation, Secure Aggregation, or client participation strategies are removed.


\subsection{\textbf{Privacy Threat Model}} \label{subsec:Privacy Threat Model}

\par In the H-FLTN framework, DERMS and EVs are considered honest-but-curious entities. While community DERMS receive all secret shares within a community, non-transitory EVs within a peer only receive partial secret shares with only 2 - 10 EVs engaging in the P2P sharing process. This ensures that no individual EV can reconstruct another EV's model weights preventing peer EVs from inferring additional information. Transitory EVs at no time participate in the P2P share process.

\par Honest-but-Curious DERMS: As aggregation centres, community DERMS receive secret shares from both transitory and non-transitory EVs. Without appropriate privacy measures, DERMS could attempt to link shared weights to specific EVs, potentially revealing sensitive mobility patterns or charging behaviours. To mitigate this concern community DERMS aggregate secret shares from both transitory and non-transitory EVs to model the global community model, and at no time reconstruct the individual local secret shares. 

\par To address these threats, the H-FLTN framework employs a multi-layered privacy strategy:

\par P2P Sharing with Augmentation: This mechanism facilitates the sharing of secret shares among non-transitory EVs. By distributing only partial information, it ensures that individual contributions remain anonymised and cannot be reconstructed by peer EVs before being securely aggregated by DERMS. This approach prevents direct associations between specific contributions and individual EVs, enhancing privacy.

\begin{assumption}
\label{assumption:honest_but_curious_derms}
We assume that community DERMS operate as honest-but-curious entities, adhering to prescribed aggregation protocols.
\end{assumption}

\par Additive Secret Sharing: ensures that community DERMS operate under privacy-preserving constraints. Non-transitory EV shares are augmented before being transmitted to the community DERMS, removing identifiable aspects before aggregation. While transitory EVs send shares directly to the DERMS, they are processed only within aggregated summations, preventing reconstruction of individual local models.
    
\par Secure Aggregation: Ensuring that only aggregated results are accessible prevents both DERMS and EVs from analysing raw weights to infer individual sensitive information.  
    
\par Secure Communications: TLS 1.3 the latest state-of-the-art cryptographic protocol adopted to secure data exchanges during communications, as discussed by Zhou et al. \cite{zhou2024challenges}. In the H-FLTN framework, TLS 1.3 is utilised for communications between DERMS and the EPDC. This protocol ensures end-to-end encryption, thereby protecting aggregated location and time predictions as they are transmitted for further analysis. Compared to its predecessors, TLS 1.3 features reduced handshake latency, streamlined encryption algorithms, and enhanced resistance to eavesdropping and tampering. By incorporating TLS 1.3, the framework guarantees secure and efficient data exchanges, maintaining the integrity of spatio-temporal predictions while minimising communication overhead.
        
\par MeLSeC: a lightweight protocol specifically designed for resource-constrained devices, as discussed in a study by Brighente et al. \cite{brighente2022melsec}, making it ideal for EV-to-peer and EV-to-DERMS communications within the H-FLTN framework. Unlike traditional protocols, which can be computationally intensive for mobile or embedded devices, MeLSeC utilises a simplified encryption mechanism that balances security and efficiency. Within the framework, MeLSeC ensures the secure transmission of sensitive EV data, such as local secret shares, while minimising energy consumption and processing delays. This tailored approach supports the efficient resource management and adaptability of the H-FLTN framework, particularly in environments with high mobility and dynamic network topologies.

\par These privacy mechanisms collectively mitigate the risks posed by honest-but-curious participants. By obfuscating individual contributions and decoupling EV identities from shared secrets, H-FLTN ensures robust privacy for all participants in dynamic and decentralised EV networks.


\section {\textbf {Discussion and Conclusion}} \label{sec:Discussion}

\par The proposed H-FLTN framework demonstrates significant advancements in predicting both the location and time of EVs’ next charging events while preserving user privacy. Location predictions achieved by the framework are highly accurate, reflecting its robustness in capturing spatial dependencies in EV mobility patterns. Additionally, the framework exhibits strong time prediction capabilities, with predictions on average falling within an hour of the actual charging event. This high level of precision enables reliable forecasting of charging times, supporting both energy demand planning and operational scheduling.

\par The framework's multi-tier hierarchical design supports decentralised collaboration among EVs, community DERMS, and the EPDC, ensuring efficient resource coordination and privacy compliance. Privacy-preserving techniques such as P2P Sharing and Augmentation, Additive Secret Sharing, Secure Aggregation, TLS 1.3, and MeLSeC were successfully integrated without significant trade-offs in model accuracy. The expanded dual-task scope of predicting both location and time introduces additional computational complexities, but the framework effectively balances these challenges while maintaining high predictive performance.

\par By providing accurate time and location predictions, the H-FLTN framework enables energy providers to accurately predict EV charging demands. Its ability to support large-scale, decentralised learning while safeguarding user privacy positions it as a practical and scalable solution for optimising energy distribution in dynamic urban environments.


\subsection{\textbf{Future Works}} \label{subsec:Future Works}

\par Building on the successes of this study and the H-FLTN framework, future research can focus on several key areas to further enhance its capabilities and computational resource management. One potential avenue is expanding the framework to predict additional metrics, such as energy consumption per charge or optimal routing patterns, which would provide energy providers with deeper insights into EV energy demands and mobility trends. Integrating these factors could support more efficient energy distribution strategies and contribute to the broader goals of sustainable energy management.

\par Additionally, improving computational efficiency in H-FLTN by refining hierarchical aggregation logic and optimising the normalisation process at the DERMS level will enhance model convergence and stability, ensuring its effectiveness in large-scale EV networks.


\section*{Acknowledgements}
The work has been supported by the Cyber Security Research Centre Limited whose activities are partially funded by the Australian Government’s Cooperative Research Centres Programme.


\bibliographystyle{IEEEtran}
\bibliography{references.bib}
\vspace{-1.6cm} 
\begin{IEEEbiography}
[{\includegraphics[width=1in,height=1.25in,clip,keepaspectratio]{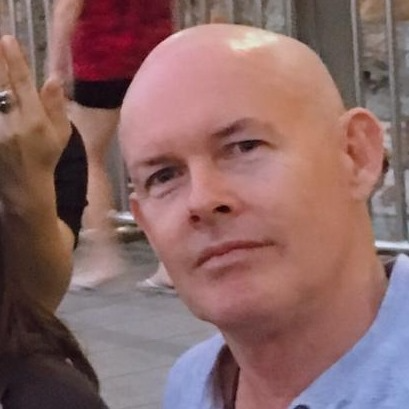}}]
{Robert Marlin}
completed a bachelor’s in information technology (BIT) at the Royal Melbourne Institute of Technology (RMIT University) in 2017, then completed a master’s in information and technology (MIT) from the Queensland University of Technology (QUT) in 2019. He worked as a Research Assistant at QUT in 2019 investigating blockchain within industry applications. Currently, he is undertaking a Ph.D. degree with QUT researching “Distributed and Privacy Preserving Analytics of Smart Grid Data”. He has published a conference paper on Advanced Topics: IoT Camera Vulnerabilities and Weaknesses, as well as co-authored a chapter on Blockchain Model for the Education Sector: A Digital Transformation Perspective of Strategic Learning, for IGI Global Press. Robert's research involves machine learning to resolve issues within the distributed energy domain, including consumer privacy and energy demand prediction.
\end{IEEEbiography}
\vspace{-.5cm} 
\vskip 0pt plus -1fil
\begin{IEEEbiography}[{\includegraphics[width=1in,height=1.25in,clip,keepaspectratio]{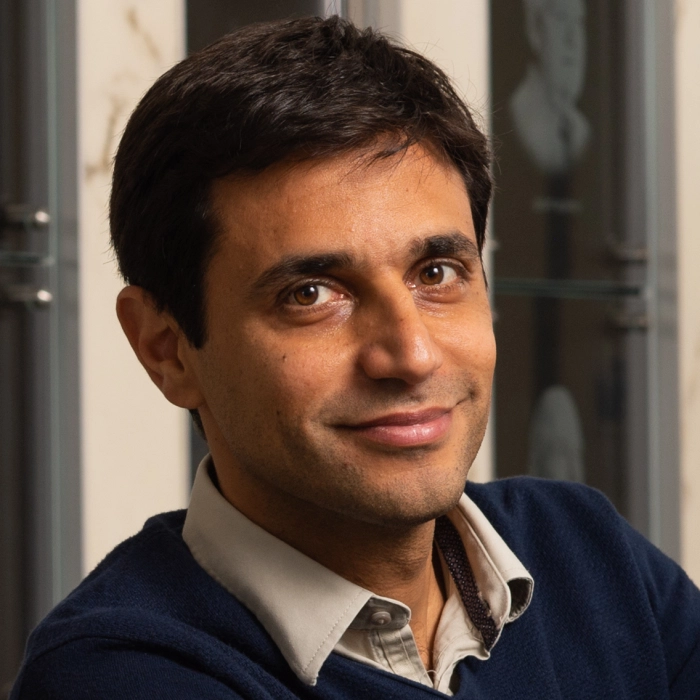}}]{Raja Jurdak}
 is a Professor of Distributed Systems and the Chair of Applied Data Sciences with the Queensland University of Technology, where he is Co-Director of the QUT Energy Transition Centre and Director of the Trusted Networks Lab. He received the M.S. and Ph.D. degrees from the University of California at Irvine. He previously established and led the Distributed Sensing Systems Group, Data61, CSIRO. He also spent time as a Visiting Academic with MIT and Oxford University in 2011 and 2017.  He has published over 280 peer-reviewed publications, including three authored books on IoT, blockchain, and cyberphysical systems. His publications have attracted over 17000 citations, with an H-index of 56. His research interests include trust, mobility, and energy efficiency in networks. He serves on the editorial boards of IEEE Transactions on Network and Service Management and Ad Hoc Networks, and on the organizing and technical program committees of top international conferences, including Percom, ICBC, IPSN, WoWMoM, and ICDCS. He is an IEEE Computer Society Distinguished Visitor, an IEEE Senior Member, and an Adjunct Professor with the University of New South Wales.
\end{IEEEbiography}
\vspace{-.5cm} 
\vskip 0pt plus -1fil
\begin{IEEEbiography}[{\includegraphics[width=1in,height=1.25in,clip,keepaspectratio]{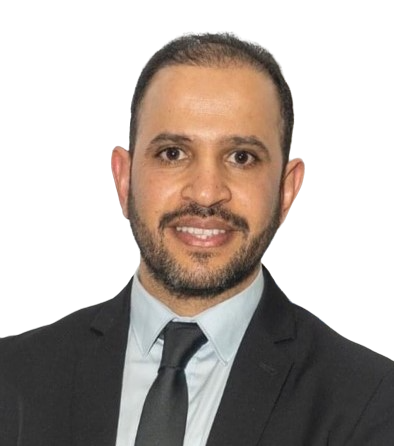}}]{Sharif Abuadbba}
is a team leader of distributed systems security at CSIRO’s Data61 . Sharif has a  Ph.D. in computer security from RMIT  University,  Australia.  He also has several years of experience working as a senior research engineer with Californian-based technology companies. He has contributions to a few US IP filled Patents in cybersecurity. He has 50+ science publications many of which in top venues including IEEE S\&P, NDSS, ACSAC, ASIACCS, ESORICS, IEEE TDSC, and IEEE TIFS. His specialist and interests include AI and cybersecurity, System and data security, and watermarking.
\end{IEEEbiography}

\end{document}